# Methodology for Building Synthetic Datasets with Virtual Humans

Shubhajit Basak
College of Engineering and Informatics
National University of Ireland, Galway
Galway, Ireland
s.basak1@nuigalway.ie

Hossein Javidnia
ADAPT Research Center
Trinity College Dublin
Dublin, Ireland
hossein.javidnia@adaptcenter.ie

Faisal Khan
College of Engineering and Informatics
National University of Ireland, Galway
Galway, Ireland
f.khan4@nuigalway.ie

Rachel McDonnell
School of Computer Science and Statistics
Trinity College Dublin
Dublin, Ireland
ramcdonn@scss.tcd.ie

Michael Schukat
College of Engineering and Informatics
National University of Ireland, Galway
Galway, Ireland
michael.schukat@nuigalway.ie

*Abstract*— **Recent advances in deep learning methods have increased the performance of face detection and recognition systems. The accuracy of these models relies on the range of variation provided in the training data. Creating a dataset that represents all variations of real-world faces is not feasible as the control over the quality of the data decreases with the size of the dataset. Repeatability of data is another challenge as it is not possible to exactly recreate 'real-world' acquisition conditions outside of the laboratory. In this work, we explore a framework to synthetically generate facial data to be used as part of a toolchain to generate very large facial datasets with a high degree of control over facial and environmental variations. Such large datasets can be used for improved, targeted training of deep neural networks. In particular, we make use of a 3D morphable face model for the rendering of multiple 2D images across a dataset of 100 synthetic identities, providing full control over image variations such as pose, illumination, and background**.

## I. Introduction

One of the main problems in modern artificial intelligence (AI) is insufficient reference data, as in many cases available datasets are too small to train Deep Neural Network (DNN) models. In some cases, where such data has been captured without a label, the manual labeling task is time-consuming, costly, and subject to human error. Producing synthetic data can be an easier approach to solving this problem. For image data, this can be achieved via three dimensional (3D) modeling tools. This approach provides the advantage of extraction of the ground truth information from 3D Computer Graphics (CG) scenes. While this process still requires some manual labor to create models, it is a one-time activity, and as a result, one can produce a potentially unlimited number of 2D pixel-perfect labeled data samples rendered from the 3D data model. The rendered data ranges from high-quality RGB images to object and class segmentation maps, accurate depth and stereo pairs from multiple camera viewpoints, point cloud data, and many more.

Generating synthetic human models including face and the full human body is even more interesting and relevant, as gathering real human datasets is more challenging than any other kind of data, mainly due to the following limitations:

- The labeling of the human face is especially complex. This includes proper head pose estimation, eye gaze detection, and facial key point detection.
- In most cases, collecting real human data falls under data privacy issues including the General Data Protection Regulation (GDPR).
- Generating 3D scans of the human body with accurate textures requires a complex and expensive full-body scanner and advanced image fusion software.
- The existing real datasets are often biased towards ethnicity, gender, race, age, or other parameters.

This synthetic data can be used for machine learning tasks in several ways:

- Synthetically generated data can be used to train the model directly and subsequently applied the model to real-world data.
- Generative models can apply domain adaptation to the synthetic data to further refine it. A common use case entails using adversarial learning to make synthetic data more realistic.
- Synthetic data can be used to augment existing real-world datasets, which reduces the bias in real data. Typically, the synthetic data will cover portions of the data distributions that are not adequately represented in a real dataset.

In this paper, we propose a pipeline using an open-source tool and a commercially available animation toolkit to generate photo-realistic human models and corresponding ground truths including RGB images and facial depth values. The proposed pipeline can be scaled to produce any number of labeled data samples by controlling the facial animations, body poses, scene illuminations, camera positions, and other scene parameters.

The rest of the paper is organized as follows: Section 2 presents a brief literature review on synthetic virtual human datasets and the motivation against this work. Section 3 explains the proposed framework. Section 4 presents some interesting results and discusses the advantages and future direction of the proposed framework.

TABLE I. REVIEW OF CURRENT SYNTHETIC VIRTUAL HUMAN DATASETS

| Dataset | 3D Model | Rigged | Full Body | 3D Background | Ground Truth |
|---|---|---|---|---|---|
| VHuF [1] | Yes | No | No | No | Facial Key points, facial Images, **No Depth Data** |
| Kortylewski et al. [3] | Yes | No | No | No | Facial Depth, Facial Images (**Only include frontal face with no Complex Background**) |
| Wang et al. [4] | Yes | No | No | No | Facial Image, Head Pose, **No depth data** |
| SyRI [5] | Yes | No | Yes | Yes | Full Body Image, **No Facial Images** |
| Chen et al. [6] | Yes | No | Yes | No | Body Pose with full body image, **No Facial Images** |
| SURREAL [7] | Yes | Yes | Yes | No | Body Pose with Image, Full Body Depth, Optical Flow, **No Facial Images** |
| Dsouza et al. [10] | Yes | No | Yes | Yes | Body Pose with Image, Depth including background, Optical Flow, **No Facial Images** |
| Ours | **Yes** | **Yes** | **Yes** | **Yes** | **Facial Images, Facial Depth including background, Head Pose** |

## II. RELATED WORK

This section presents an overview of existing 3D virtual human datasets and their applications. It also describes their limitations, which are the main motivation of this work.

Queiroz et al. [1] first introduced a pipeline to generate facial ground truth with synthetic faces using the FaceGen Modeller [2], which uses morphable models to get realistic face skin textures from real human photos. Their work resulted in a dataset called Virtual Human Faces Database (VHuF). VHuF does not contain the ground truth like depth, optical flow, scene illumination details, head pose, and it only contains head models that are not rigged and placed in front of an image as a background. Similarly, Kortylewski et al. [3] proposed a pipeline to create synthetic faces based on the 3D Morphable Model (3DMM) and Basel Face Model (BFM-2017). They only captured the head pose and facial depth by placing the head mesh in the 2D background. The models are not rigged as well. Wang et al. [4] introduced a rendering pipeline to synthesize head images and their corresponding head poses using FaceGen to create the head models and Unity 3D to render images, but they only captured head pose as the ground truth and there is no background. Bak et al. [5] presented the dataset Synthetic Data for person Re-Identification (SyRI), which uses Adobe Fuse CC for 3D scans of real humans and the Unreal Engine 4 for real-time rendering. They used the rendering engine to create different realistic illumination conditions including indoor and outdoor scenes and introduce a novel domain adaptation method that uses synthetic data.

Another common use case of virtual human models is in human action recognition and pose estimation. Chen et al. [6] generated large-scale synthetic images from 3D models and transferred the clothing textures from real images, to predict pose with Convolution Neural Networks (CNN). It only captured the Body Pose as the ground truth. Varol et al. [7] introduced the SURREAL (Synthetic hUmans foR REAL tasks) dataset with 6 million frames with ground truth pose, the depth map, and a segmentation map that showed promising results on accurate human depth estimation and human part segmentation in real RGB images. They used the SMPL [8] (Skinned Multi-Person Linear) body model trained on the CAESAR dataset [9], one of the largest commercially available data that has 3D scans of over 4500 American and European subjects, to learn the body shape and textures, CMU MoCap to learn the body pose, and Blender to render and accumulate ground truth with different lighting conditions and camera models. Though this is the closest work to this paper that can be found, the human models are not placed in the 3D background, instead, they are rendered using a background image. It also did not capture the Facial Ground Truths as it focused on the full-body pose and optical flow. Dsouza et al. [10] introduced a synthetic video dataset of virtual humans PHAV (Procedural Human Action Videos) that also uses a game engine to obtain the ground truth like RGB images, semantic and instance segmentation, the depth map, and optical flow, but it also does not capture Human Facial Ground truths.

Though there are previous works on creating synthetic indoor-outdoor scenes and other 3D objects, there is limited work done on exploring the existing available open-source tools and other commercially available software to build a large dataset of synthetic human models. Also, another major concern is the realism of the data and per-pixel ground truth. The proposed method tries to fill that gap. It can generate realistic human face data with 3D background and capturing the ground truths like head pose, depth, optical flow, and other segmentation data. As these are fully rigged full-body models, body pose with the other ground truths can also be captured. A detailed featurewise comparison can be found in table 1.

## III. METHODOLOGY

This section presents a detailed framework for generating the synthetic dataset including RGB images and the corresponding ground truth.

### A. 3D Virtual Humans and Facial Animations

The iClone 7 [11] and the Character Creator [12] software is used to create virtual human models. The major advantages of using iClone and Character Creator are:

- Character Creator provides "Realistic Human 100" models that reduce the bias over ethnicity, race, gender, and age. These pre-built templates can be applied to the base body template as shown in Fig. 1.

- The morphing of different parts of the body can be adjusted to create more variations to the model. Fig. 2 shows adjustment in cheek, forehead, skull, and chin bone.

This work is funded by Science Foundation Ireland Centre for Research Training in Digitally Enhanced Reality (D-REAL) under grant 18/CRT/6224.

- Different expressions including neutral, sad, angry, happy, and scared can be added to the models to create facial variations. Fig. 3 presents a sample render of these five expressions from iClone.

- The models provide Physically Based Rendering (PBR) textures (Diffuse, Opacity, Metallic, Roughness) to render high-quality images.

- Models can be exported in different formats (like obj, fbx, and alembic) which are supported by the most popular rendering engines.

Though iClone can render high-quality images, it does not provide the functionality to capture other ground truth data like exact camera locations, head pose, scene illumination details. Therefore, the models were exported from iClone and placed in a 3D scene in the popular free and open-source 3D CG software toolset Blender [13]

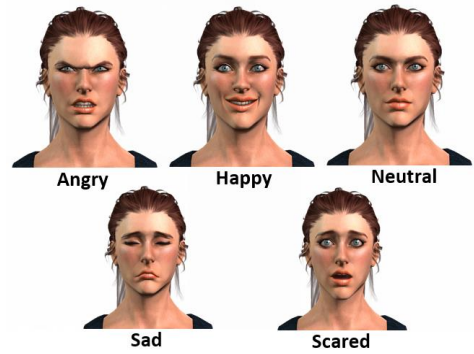

Fig. 3. Sample images with different expression rendered from iClone

In this research, the FBX format is used as it exports the model with proper rigging, which helps to add movements to different body parts including the head. A sample of a fully rigged model is shown in Fig. 4 after the model is loaded in Blender.

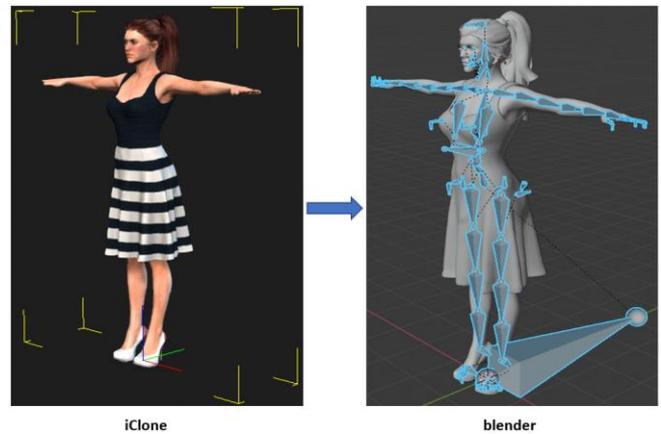

Fig. 4. Sample of a fully rigged model imported in Blender from iClone

### C. Rendering

The iClone models are imported to Blender 3D modeling software.

The major components of Blender are Models, Textures, Lighting, Animations, Camera Control (including lens selection, image size, focal length, the field of view (FOV), movement, and tracking), and the rendering engine. The two most common and popular render engines supported by Blender are Cycles and Eevee. Cycles uses a method called path tracing, which follows the path of light and considers reflection, refraction, and absorption to get the realistic rendering, while Eevee uses a method called rasterization, which works with the pixel information instead of paths of light, which makes it fast but reduces the accuracy. A good comparison of these two rendering engines can be found in [14]. A sample workflow of the major components of Blender is described in Fig. 5.

In the current work the following steps are taken to obtain the final output:

- To replicate the process of capturing real data, the camera is placed at a fixed location in the scene and the relative distance from the model to the camera center is varied within a range of 700 mm to 1000 mm to the human model as shown in Fig. 6.

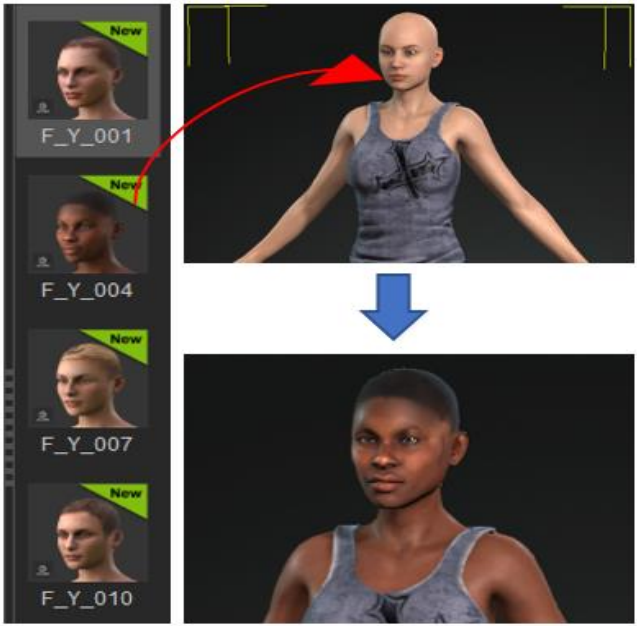

Fig. 1. Applying head template on a base female template in Character Creator

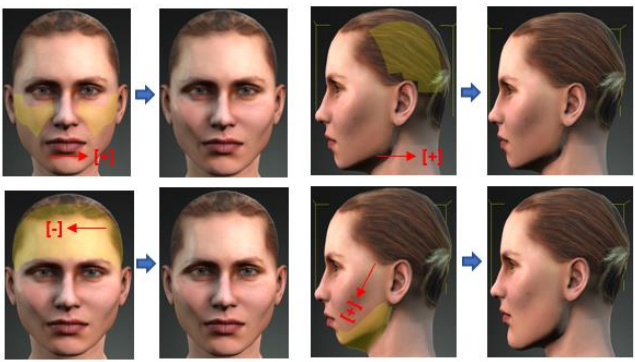

Fig. 2. Adjust cheek, forehead, skull and chin bones in Character Creator

### B. Model Exporting from iClone

The model created in iClone can be exported in different formats that are supported by the most popular 3D modeling software including Blender. Two of these formats are explored in this work including Alembic (.abc) and FBX (.fbx).

- Different illumination is added to the 3D scene which can be varied to create different realistic lighting which includes point, sun, spotlight, and area light.

- Different render passes are set up in Blender to get the RGB and the corresponding depth images. Cycles rendering engine is used to get a realistic rendering. It has been observed during the rendering of the transparent materials that Cycles path tracing can cause noisy output. To reduce the noise, the branched path tracing is used. It splits the path of the ray as the ray hits the surface and takes into account the light from multiple directions and provide more control for different shaders.

- As the model is rigged, the movement of most of the body parts can be controlled by selecting their bone structure. Here the shoulder and head bones are selected, and the head mesh is rotated with respect to those bones.

Rotations of yaw (+30 degree to -30 degree), roll (+15 degree to -15 degree), and pitch (+15 degree to -15 degree) are applied to the head and the keyframes are saved. Later these keyframes are used to capture the head pose. A sample setup in Blender is illustrated in Fig. 7.

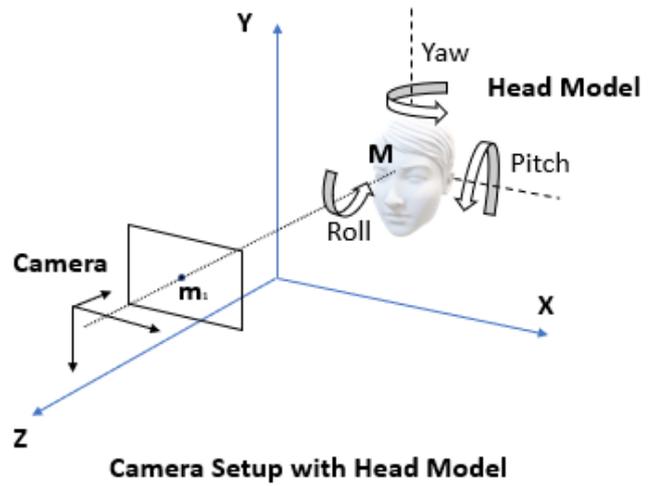

Fig. 6. Sample setup of camera and the model

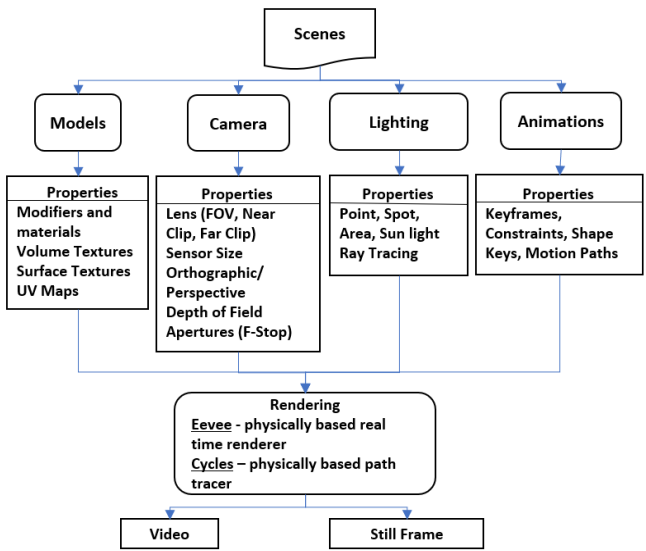

Fig. 5. Sample workflow in Blender

Following the above three steps, the proposed framework works as follows: Using the Real 100 head models a set of virtual human models is created in Character Creator. The texture and morphology of the models are modified to introduce more variations. These models are then sent to iClone where five facial expressions are imposed. The final iClone models with the facial expressions are exported in FBX which consists of the mesh, textures, and animation keyframes.

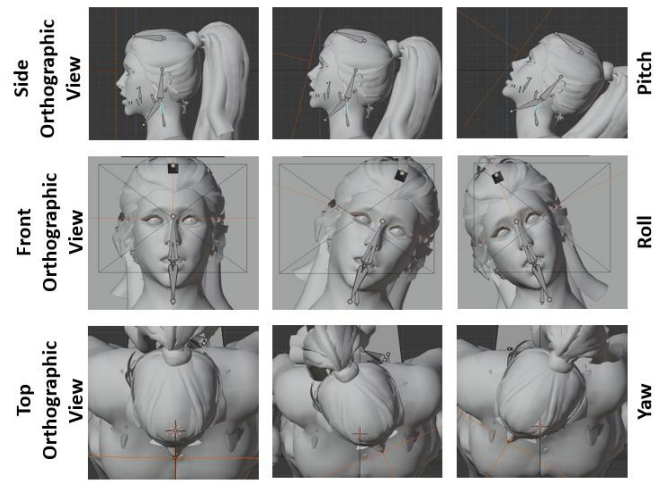

Fig. 7. Applying head movement (yaw, roll, and pitch) on the model in Blender to capture the head pose

The FBX files are then imported and scaled in the Blender world coordinate system. Lights and cameras are added to the scene, whose properties are then adjusted to replicate the real environment. The near and far clip of the camera is set to 0.01 meters and 5 meters respectively. The FOV and the camera sensor size are set to 60 degrees and 36 millimeters respectively. The RGB and Z-pass output of the render layer is then set up in the compositor to get the final result. To apply the rotation, the head and shoulder bone is identified in pose mode and the head mesh is rotated with respect to those bones, and the keyframes are saved. Finally the all the keyframes are rendered to get the RGB and the depth images and the respective head pose (yaw, pitch, and roll) is captured through the python plugin provided by Blender. The overall pipeline is described in Fig. 8.

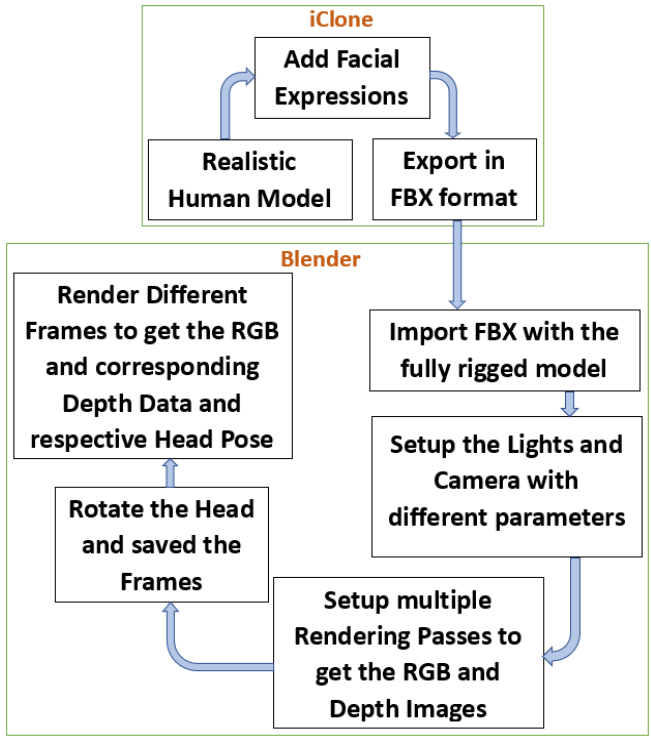

Fig. 8. Pipeline to produce a virtual human

## IV. RESULTS AND DISCUSSIONS

Using the framework proposed in Section III, several virtual human models with their corresponding RGB and depth images have been rendered.

The experiments and data generation is performed on an Intel Core i5-7400 3 GHz CPU with 32 GB of RAM equipped with an NVIDIA GeForce GTX TITAN X Graphical Processing Unit (GPU) having 12 GB of dedicated graphics memory. The RGB and depth images are rendered with a resolution of 640 X 480 pixels and their raw depth is saved in .exr format. The average rendering time for each frame is 57.6 seconds. The models are rendered in Blender using different parameters such as the positions of lights, camera parameters, keyframe values of the saved animations. The raw binary depth information and the head pose information are also captured as part of this dataset. Fig. 9 presents the RGB images and their corresponding ground truth depth images (scaled to visualize) with a different head pose. Fig. 10 shows the results with different illuminations. The models then imported to more complex 3D scenes and the ground truth data has been captured. Fig. 11 shows some samples and the corresponding depth with complex backgrounds.

The proposed method allows the creation of potentially unlimited data samples with pixel-perfect ground truth data from the 3D models. Also, the 3D models can be placed in any 3D scene and the data can be rendered within a different environment. Another advantage of using this pipeline of tools is that the positions of the camera and their intrinsic parameters and the scene lighting can be controlled to replicate a real environment. As these models have PBR shading and blender cycle rendering engine utilizes the path ray tracing and accurate bounce lighting the rendered images are more realistic than the previous datasets present. Table 2 provides some samples from other datasets that capture facial synthetic data and shows the result from the proposed model is more realistic and robust than the previous ones. Although the proposed pipeline can generate a large amount of data more work has to be done in domain transfer and domain adaptation areas to make the images as realistic as possible.

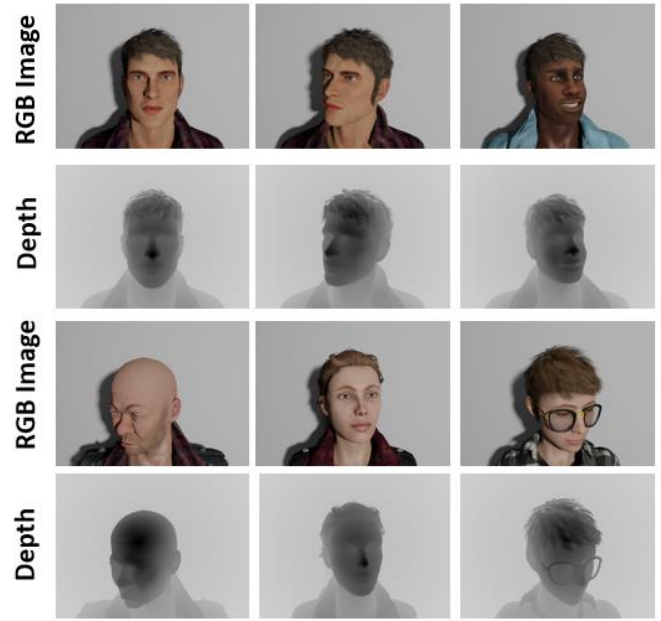

Fig. 9. Sample images of virtual human faces and their ground truth depth (scaled to visualize) with different head pose

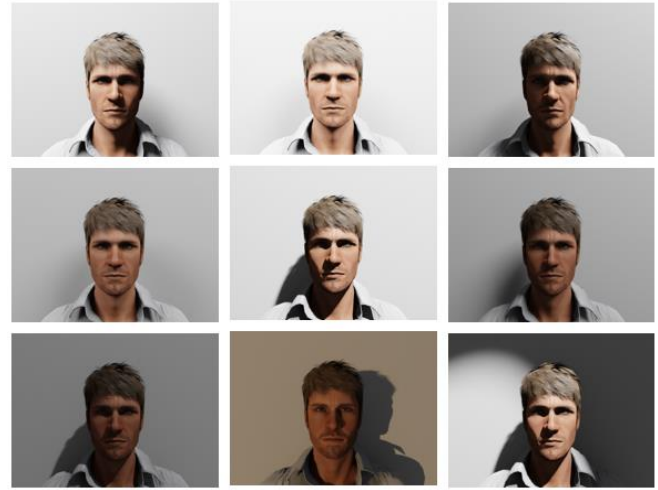

Fig. 10. Sample images of virtual human faces in different lighting condition

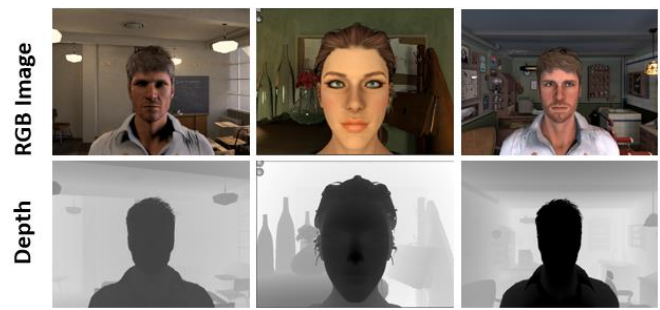

Fig. 11. Sample images and their depth image (scaled to visualize) with more complex background

TABLE II.   IMAGE SAMPLES FROM EXISTING FACIAL SYNTHETIC DATASET

| Dataset | Ground Truth |
|---|---|
| VHuF [1] | 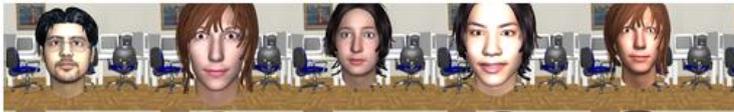 |
| Kortylewski et al. [3] | 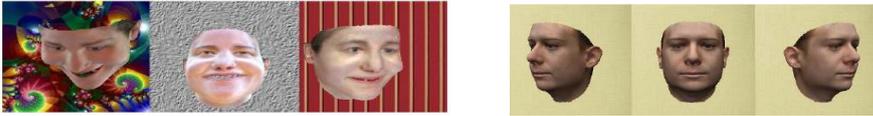 |
| Wang et al. [4] | 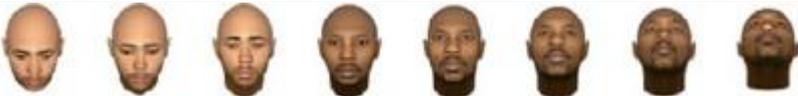 |
| Ours | 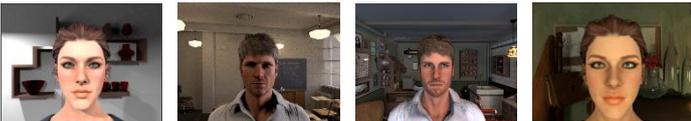 |

## V. CONCLUSION

In this work, a framework to synthetically generate a huge set of facial data with variations in environment and facial expressions using available toolchains is explored. This will help to train DNN models, as it covers more variations in expressions and identity. Previously generated synthetic human datasets [6], [7] mostly lack realism and per-pixel ground truth data. The proposed pipeline will help to overcome such limitations. The data generated through this framework can extensively be used for facial depth estimation problems. There are currently a few datasets available with real-world facial images and their corresponding depth [15],[16],[17],[18]. However, it is practically impossible to get pixel-perfect depth images of the human faces due to the limitation of the available sensors like Kinect. The proposed framework can bridge this gap with more accurate ground truth facial depth data. The models can also be used to build more advanced 3D scenes which will cover more complex computer vision tasks such as driver monitoring system, 3D aided face recognition, elderly care, and monitoring.

## ACKNOWLEDGMENT

This material is based upon works supported by the Science Foundation Ireland Centre for Research Training in Digitally Enhanced Reality (D-REAL) under grant 18/CRT/6224.